\documentclass[english]{article}
\usepackage[T1]{fontenc}
\usepackage[latin9]{inputenc}
\usepackage{geometry}
\usepackage{url}
\usepackage{amsmath}
\usepackage{graphicx}

\makeatletter

\providecommand{\tabularnewline}{\\}

\usepackage{balance} 
\usepackage{marginnote}
\usepackage{geometry}
\date{}
\usepackage{newtxtext,newtxmath}

\@ifundefined{showcaptionsetup}{}{%
 \PassOptionsToPackage{caption=false}{subfig}}
\usepackage{subfig}
\makeatother

\usepackage{babel}
\begin{document}
\title{Artificial mental phenomena: Psychophysics as a framework to detect
perception biases in AI models}
\author{Lizhen Liang and Daniel E. Acuna\thanks{Corresponding author: \protect\url{deacuna@syr.edu}}\\
School of Information Studies\\
Syracuse University\\
Syracuse, New York}
\maketitle
\begin{abstract}
Detecting biases in artificial intelligence has become difficult because
of the impenetrable nature of deep learning. The central difficulty
is in relating unobservable phenomena deep inside models with observable,
outside quantities that we can measure from inputs and outputs. For
example, can we detect gendered perceptions of occupations (e.g.,
female librarian, male electrician) using questions to and answers
from a word embedding-based system? Current techniques for detecting
biases are often customized for a task, dataset, or method, affecting
their generalization. In this work, we draw from Psychophysics in
Experimental Psychology---meant to relate quantities from the real
world (i.e., ``Physics'') into subjective measures in the mind (i.e.,
``Psyche'')---to propose an intellectually coherent and generalizable
framework to detect biases in AI. Specifically, we adapt the two-alternative
forced choice task (2AFC) to estimate potential biases and the strength
of those biases in black-box models. We successfully reproduce previously-known
biased perceptions in word embeddings and sentiment analysis predictions.
We discuss how concepts in experimental psychology can be naturally
applied to understanding artificial mental phenomena, and how psychophysics
can form a useful methodological foundation to study fairness in AI.
\end{abstract}

\section{Introduction}

Recent artificial intelligence models have shown remarkable performance
in a variety of tasks that were once thought to be solvable only by
humans \cite{stone2016one}. With such promising results, companies
and governments begun deploying such systems for increasingly critical
tasks, including job candidate screening \cite{dastin2018amazon},
justice system decisions \cite{angwin2016machine}, and credit scoring
\cite{hurley2016credit}. Due to training with data that might contain
biases, however, deep learning models inadvertently fit those biases
and create decisions that discriminate against gender and other protected
statuses. If we were to find those biases in humans, we could interrogate
them and determine whether such biases have occurred. Several researchers
have attempted to develop methods for detecting biases in AI models,
but these methods are specific to the task (e.g., \cite{stock2017convnets}),
data (see \cite{chen2018my}), or type of model \cite{celis2019classification}---hindering
their potential adoption. Here we entertain the idea of using Experimental
Psychology to develop novel and coherent methods for probing AI systems.
Experimental Psychology has a very rich tradition of treating human
consciousness as a black box, developing and extracting potential
biases from subjective judgements in behavioral tasks \cite{fechner}.
We hypothesize that we can adapt these methods to uncover biases in
AI models in a similar way. In particular, Psychophysics and signal
detection theory offer a concrete set of tools for querying black
boxes and extracting useful measures on the direction and strength
of bias. In this work, we describe how we adapt the standard two-alternative
forced choice (2AFC) task, the workhorse of Psychophysics, to extract
biases in word embeddings and sentiment analysis predictions.

The dramatic increase in the usage of AI systems has called into question
potential biases made against vulnerable groups. Part of the issue
is that current systems have exploded in their complexity, going from
hundreds of parameters linearly related to outputs, to billions of
parameters non-trivially related to outputs (as discussed broadly
in \cite{hastie2005elements}, \cite{tan2019efficientnet}, and \cite{o2016weapons}).
If biases are present in modern AI systems, they are therefore significantly
harder to detect just by inspecting fitted parameters. This has resulted
in dramatic cases of discrimination in recidivism prediction \cite{angwin2016machine}
and credit scoring \cite{hurley2016credit}, which are only discovered
once systems are deployed. One proposed solution to the problem of
veiled discrimination, recently implemented in Europe's General Data
Protection Regulation (GDPR), is to force AI models to be ``explainable''
\cite{dovsilovic2018explainable}. While forcing explainability appears
as a natural countermeasure, the well-known interpretability--accuracy
trade-off would predict that these systems have decreased accuracy
\cite{sarkar2016accuracy}. This is not always desirable \cite{domingos2015master}.

One solution for preventing biases present in AI models is to develop
techniques for detecting them, as a natural first step to fixing them.
There have been several research programs aimed at detecting biases
in AI models. Many of them, however, require detailed knowledge of
the inner workings of the algorithm or the datasets. For example,
in the work of \cite{mcduff2019characterizing}, the authors propose
a form of ``classifier interrogation'' which requires labeled data
to explore the space of parameters that might cause biases. Also,
techniques for detecting biases are somewhat task specific and difficult
to generalize. In \cite{caliskan2017semantics}, for example, the
authors adapt the Implicit Association Task (IAT) for detecting biases
in word embeddings. While this is a natural application of the original
intention of IAT, it is unclear how to move beyond bias detection
in unsupervised settings. Recently, researchers in DeepMind proposed
Psychlab, a highly-complex synthetic environment to test AI models
as if they were humans living in a virtual world \cite{leibo2018psychlab}.
Detecting biases of AI models is an important step but it would be
beneficial to develop more general-purpose and simpler techniques.

Interestingly, Experimental Psychology has had to develop methods
for understanding latent, mental phenomena based on questions and
answers that are exerted verbally or physically. In particular, the
field of Psychophysics uses methods to measure the perception biases
and sense's accuracies of animals \cite{green1966signal}. Importantly,
a key requirement of Psychophysics is to avoid relying on verbal or
other highly-cognitive responses that are prone to noise and rather
use simple behavioral responses that are hard to fake (e.g., movements,
yes/no answers). In the whole of Psychophysics, perhaps one of the
oldest and most well-developed techniques for performing estimations
based on these cues is the two-alternative forced choice task (2AFC)
\cite{acuna2015using}. This method is used, for example, to measure
how the size of objects biases our perception of weight \cite{charpentier1891analyse},
and to measure the precision of the human retina when detecting light
\cite{harmening2014mapping}. We hypothesize that we can adapt these
techniques for measuring biases and the strength of those biases in
AI models. Thus, Experimental Psychology is a rich area of research
with potential applications to examine artificial intelligence decisions.

In this work, we develop a framework to study biases in AI models.
Our primary goal is to develop a framework that is coherent across
datasets, tasks, and algorithms, allowing a researcher to describe
biased perception using a common language. We draw inspiration from
Psychophysics, a field of Experimental Psychology, and adapt the two-alternative
forced choice (2AFC) task. As an example, we examine potential biases
in word embeddings and sentiment analysis predictions and we validate
our results with real-world data. Our findings show that we are able
to detect biases and the strength of them in decisions that involve
gender and occupations. In sum, our work contributes to the field
of fairness, accountability, and transparency in the following manner:
\begin{itemize}
\item A discussion of the current bias detection techniques for AI models
\item A new method for detecting biases and measuring the strength of those
biases based on a coherent set of concepts and language drawn from
Experimental Psychology
\item A demonstration of the application of the technique to word embeddings
and sentiment analysis prediction
\end{itemize}

\section{Background}

\subsection{Different Kinds of Biases}

\textbf{Psychophysical\ bias} Perceptual biases are decision deviations
about stimuli that should be perceived exactly the same. A classic
example is the size-weight illusion (known as the Charpentier illusion)
in which people underestimate the weight of a large object compared
to small objects of the same mass \cite{charpentier1891analyse}.
Formulated differently, if presented with a small object of a known
weight, subjects would tend to judge objects of larger size to have
the same weight as the small object. Therefore, even though the known
small object has the same size of another object of unknown weight,
subjects would not judge the other unknown-weight object as having
the same weight: they would have a bias against small sizes in weight
space. This is the kind of bias that we are expecting to detect with
the techniques introduced here.

\textbf{Discriminatory\ bias/Anti-discrimination\ laws} Discrimination
towards gender, race, sex, and ethnicity is generally considered a
violation to the fourteenth amendment \cite{bornstein2018antidiscriminatory}.
This kind of bias lives in the judiciary system where laws and acts
have been designed to protect the rights of certain groups of people.
Some AI models being utilized by companies, governments and other
organizations may inherit discriminatory biases that are unlawful
in this sense \cite{o2016weapons}. This is the kind of bias we want
to help detect by adapting the Psychophysical experiments explained
above. However, there must be a human judge or an external validation
of whether these detected biases go against, for example, anti-discrimination
laws \cite{benthall2019racial}.

\textbf{Statistical\ Biases} This bias represents the difference
between an estimated data distribution and a real data distribution
\cite{hastie2005elements}. A statistical model with low bias indicates
that the model has low training error but overfits and performs poorly
on testing (e.g., out-of-sample) data. In this context, bias might
help prevent overfitting by forcing the model not to fit the data
too well. Several techniques in machine learning and statistical modeling
(such as prior probabilities, regularization, dropout, and data augmentation
\cite{murphy2012machine}) are meant to introduce bias in the system
with the purpose of preventing overfitting.

\textbf{Counterfactual\ Biases} In this kind of bias, the question
is more specific: would the response of the system change had one
protected attribute in the input been different? In this instance,
of course, it is not possible to go back in time and change the situation,
and thus many assumptions must be made. With the mathematical framework
of causal reasoning (e.g., \cite{pearl2009causality,imbens2015causal}),
researchers have proposed ways to use counterfactual reasoning to
think about these issues (e.g., \cite{kusner2017counterfactual,kilbertus2019sensitivity,wu2019counterfactual}).
Counterfactual biases are orthogonal and complementary to psychophysical
biases. In general, Psychophysics does not deal with causal inference
because it is believed that the experimenter controls for potential
confounders or relies on randomization to assign subjects to experimental
conditions.

\begin{figure*}
\centering{}\includegraphics[width=1\textwidth]{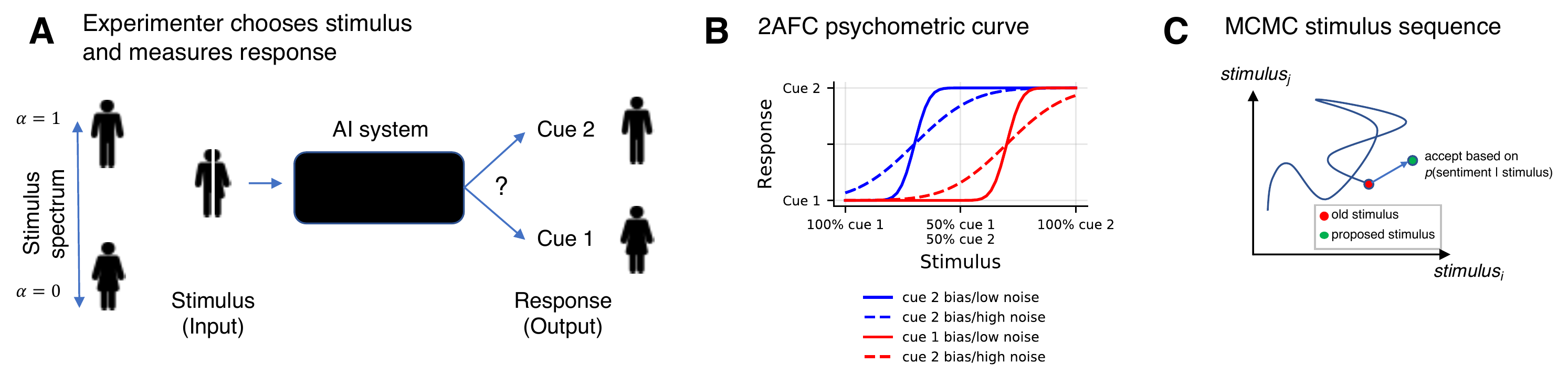}\caption{\label{fig:Artificial-Psychophysics:-two-al}Artificial Psychophysics:
two-alternative forced choice task (2AFC) \textbf{A)} experimenter
presents a stimulus that is a blend of two cues. The mixture amount
is controlled by a parameter $\alpha$. The AI system responds with
one of two choices. \textbf{B) }A psychometric curve that best fits
a set of responses to several $\alpha$ values. The point at which
the curve crosses the 50/50 $y$-axis is called the Point of Subjective
Equivalence (PSE) and the \emph{slope} of such curve is inversely-related
to the Just Noticeable Difference (JND). \textbf{C}) We can use the
responses of the system to create a sampler based on Markov Chain
Monte Carlo (MCMC) to extract the distribution of stimuli that produces
a certain output. The panel shows an example where the sampler uses
responses to estimate stimulus (e.g., embeddings) of positive sentiments.}
\end{figure*}

\subsection{Psychophysics and the two-alternative forced choice task (2AFC)}

Psychophysics is perhaps one of the oldest parts of Psychology, established
in a book by Fechner in 1860 \cite{fechner}. It emphasizes the quantification
of the relationship between physical stimulus (light, sound, touch)
with the contents of ``consciousness'', which are unmeasurable (This
stands in contrast to other approaches based on behavior and verbal
interviews \cite{murray1993perspective}.) Psychophysics had an early
influence from sophisticated mathematical tools rooted in signal detection
theory---a field that seeks to model responses of systems based on
a mathematical/statistical treatment of signals \cite{green1966signal}.
One of the simplest and perhaps most widely used methods in Psychophysics
is the two-alternative forced choice (2AFC) task. In this task, subjects
are asked to repeatedly answer one of two questions based on stimuli
carefully selected by the experimenter. Depending on the set of answers,
then, Psychophysics uses curve fitting and interpretation to extract
perception biases. This task has been used to establish many important
findings about our memory, visual, and auditory systems \cite{green1966signal}.
Thus, Psychophysics is one of the oldest branches of Psychology and
2AFC is a workhorse paradigm, widely used to measure biases.

In the 2AFC task, there are two important quantities that are obtained
from a repeated set of questions. One quantity is the Point of Subjective
Equivalence (PSE). As its name indicates, in the 2AFC task, a subject
is ``forced'' to make one of two choices about a stimulus even if
neither of the answers seems correct. The stimulus at which this extreme
confusion happens is called the Point of Subjective Equivalence (PSE)
because both choices are ``subjectively'' similar. Another quantity
is the Just-Noticeable Difference (JND) which is the amount of stimulus
that the experiment needs to modify in order for the subject to reliably
shift answers. If the PSE is not where the experimenter expects it
to be, then we say that the subject has a bias \cite{green1966signal}.
This entire process of querying and estimating PSE and JNE can be
observed in Fig. \ref{fig:Artificial-Psychophysics:-two-al}A. There,
the experimenter presents a stimulus selected from an spectrum of
choices. This stimulus is then taken as an input by the subject (AI
system), who is asked to decide whether one of two cues was used to
generate the stimulus. Based on many of these responses, a \emph{psychometric}
curve is fit using standard cumulative functions such as the sigmoid
or cumulative normal distribution (Fig. \ref{fig:Artificial-Psychophysics:-two-al}B).
The point at which this psychometric curve passes through the 50/50
is the PSE. The JND is related to the inverse of the steepness of
such curve.

Mathematically, and without loss of generality, we can assume that
the subject (or AI system) is presented with a stimulus $s$ selected
from a stimulus spectrum. This stimulus in turn is a combination (e.g.,
linear) of two cues $c_{1}$ and $c_{2}$ using a parameter $\alpha\in[0,1]$,
as follows
\begin{equation}
s=(1-\alpha)c_{1}+\alpha c_{2}.\label{eq:stimulus}
\end{equation}

The subject only perceives a noisy version of Eq. \ref{eq:stimulus}
denoted by $s'$. The subject has a prior on the general values of
cue 1 and 2, $p(c_{1})$ and $p(c_{2})$, respectively. Also, the
subject has a general idea of the perception of the stimulus, $s_{p}$,
given a hypothesized stimulus, $\hat{s}$, $p(s_{p}\mid\hat{s})$.
With all these pieces of information, the subject can estimate the
distribution of the real stimulus, $\hat{s}$, given the perceived
stimulus using Bayes' rule
\begin{equation}
p(\hat{s}\mid s_{p})=\frac{p(s_{p}\mid\hat{s})p(\hat{s})}{p(s_{p})}.\label{eq:posterior}
\end{equation}

Based on Eq. \ref{eq:posterior}, and using some scoring function
$\text{score}(s,c)$ relating the stimulus $s$ with cue $c$, the
decision of the subject, $\omega$, for a given perception $s_{p}$
and a hypothesized stimulus $\hat{s}$ is
\begin{equation}
\omega(\hat{s})=\begin{cases}
\text{cue 1} & \text{score}(\hat{s},c_{1})>\text{score}(\hat{s},c_{2})\\
\text{cue 2} & \text{o.w.}
\end{cases}\label{eq:decision-function}
\end{equation}

Because there is noise in the perception (i.e., $p(s_{p}\mid\hat{s})$),
then this decision might change from trial to trial for a given stimulus
$s$. This is largely similar to the Bayesian treatment of the 2AFC
task (see \cite{acuna2015using,berniker2010learning,kording2004bayesian,ernst2002humans,yuille1993bayesian,knill1996perception}).
As it generally assumed that there is no bias in $\hat{s}$ with respect
to the real $s$, the hypothesized stimulus $\hat{s}$ (or $s$) is
largely determined through the mixture $\alpha$. A function that
produces the probability of whether to pick $c_{2}$ over $c_{1}$
is called a \emph{psychometric curve} ($\Psi$) and it is defined
as follows
\begin{align}
\Psi(\alpha) & =\iiiint p(\omega(\hat{s})=\text{"cue 2"},c_{1},c_{2},s_{p},\hat{s})\thinspace\text{d}c_{1}\thinspace\text{d}c_{2}\thinspace\text{d}s_{p}\text{d}_{\hat{s}}.\label{eq:psych-curve-1}
\end{align}
This psychometric curve is usually assumed to be a cumulative distribution
function and thus monotonically increasing in $\alpha$.

In this context, then, the Point of Subjective Equivalence (PSE) can
be defined as the value of $\alpha$ for which the psychometric curve
has a 50/50 chance of answering either cue 1 or 2
\begin{equation}
\text{PSE}=\arg\text{\text{solve}}_{\alpha}\Psi(\alpha)=\frac{1}{2}.\label{eq:PSE}
\end{equation}

And the Just-Noticeable Difference (JND) is the amount of $\alpha$
where there is a noticeable difference in the decisions in the psychometric
curve of say 50\%
\begin{equation}
\text{JND}=\arg\text{solve}_{\Delta\alpha}\Psi(\text{PSE}+\frac{\text{\ensuremath{\Delta\alpha}}}{2})-\Psi(\text{PSE}-\frac{\text{\ensuremath{\Delta\alpha}}}{2})=\frac{1}{2}.\label{eq:JND}
\end{equation}

One of our interests in this study is to understand biases in the
perception of cues. If there were no biases, it would be expected
that the PSE is $1/2$ because a mixture of $\alpha=1/2$ should make
the stimulus equally similar to cue 1 and cue 2. However, this is
not always the case. A PSE > 1/2 can be interpreted as a bias \emph{for
}cue 1 (or \emph{against }cue 2) as a higher than 50\% proportion
of cue 2 (and lower than 50\% proportion of cue 1) would be needed
to make the cues perceptually indistinguishable. Conversely, a PSE
< 1/2 can be interpreted as a bias \emph{for }cue 2 (or \emph{against}
cue 1). The value of JND depends on the perception noise of the task.
A large JND means that perceptions are noisy and biases (if any) are
less sharply defined. It is typically assumed that there is no correlation
between PSE and JND.

\subsection{Markov Chain Monte Carlo (MCMC) for stimulus representations}

While the 2AFC task allows to measure biases and the strength of those
biases based on stimulus chosen by the experimenter, it would be desirable
to reverse the process. This is, it would be interesting to understand
the distribution of the stimulus for a given response. In a classification
task, for example, this would be useful to understand the distribution
of the texts that give rise to positive sentiment predictions. This
idea has been explored before in the context of psychological experiments
\cite{sanborn2010uncovering} by using a specific type of Markov Chain
Monte Carlo (MCMC) sampler. Because in our experiment we can control
how we treat the probabilistic outcome of a classifier, we can use
a highly-efficient MCMC method such as the No-U-Turn sampler \cite{hoffman2014no}.

Concretely, imagine that we want to understand how the input of a
classifier, $s$, is related to its decision, $\omega$. Without loss
of generality, we assume that we have access to the classifier's distribution
on $\omega$ given the input $s$ as $p(\omega\mid s)$. We reverse
this distribution by simply applying Bayes' rule
\begin{equation}
p(s\mid\omega)=\frac{p(\omega\mid s)p(\omega)}{p(\omega)}.\label{eq:bayes-rule}
\end{equation}

When the dimensionality of the input $s$ is high, such as in most
modern deep learning applications, estimating $p(\omega)$ is prohibitively
expensive because we need to integrate out all dimensions of $s$
from the join distribution $p(\omega,s)$. Therefore, we can use a
Markov Chain Monte Carlo (MCMC) scheme where, in a similar fashion
to the 2AFC task, we repeatedly ask the system for its judgements
about an input. In our context, a sampler like this, being at a certain
embedding $s_{t}$ attempts to move to another embedding $s_{t}'$
which is only accepted if the MCMC acceptance function is met (Fig.
\ref{fig:Artificial-Psychophysics:-two-al}C). For more information
on MCMC, see \cite{gilks1995markov}.

\section{Proposed method}

\subsection{Estimating bias and bias strength in word embeddings using the 2AFC
task}

Based on artificial neural networks, word embedding models compute
a continuous representation of a word using contextual word co-occurrences
within documents. These representations work especially well for language
translation and word analogy tasks \cite{schnabel2015evaluation}.
For our experiments, we use a word embedding method called GloVe \cite{pennington2014glove}
but we believe any other embedding method should produce similar results.

To examine potential biases in word embeddings, we design an artificial
2AFC task where a system is asked to answer questions about a concept
that should be unbiased as it relates to two potentially biasing concepts.
Consider a real 2AFC task examining genderless occupations and their
relationship to genders. For example, we can ask participants to guess
the gender of an electrician---e.g., a person with a voltage meter
and a blue coat---whose face and body have been experimentally manipulated
to be a blend between a male and a female face. By modifying the percent
of maleness blending, we would obtain a psychometric curve based on
responses. If such psychometric curve crosses the 50/50 threshold
away from a 50/50 gender blending, it would suggest a biased perception
of the occupation. It is worth mentioning that this experiment would
be challenging to perform in humans because of inter-trial memory
effects and because the visual blending of faces and body needs to
be believable. With an AI model, however, these issues are not present.

We need to create an artificial 2AFC task with word embeddings. We
use a simple question--answering system solely based on distances.
In word embeddings, close relationships between words are well correlated
with the angles between their respective embeddings (i.e., cosine
distances). The adaptation of the task explained above would be as
follows: we would ask the AI system about the gender in the question
``What is the gender of this {[}\emph{manipulated gendered pronoun}{]}
electrician?'' with the answers being a female attribute or male
attribute (e.g., female/male, woman/man, her/him). The manipulated
gender pronoun would be the stimulus and ``electrician'' would be
the occupation of interest. The stimulus would be represented by a
mixture of a female attribute embedding $c_{1}$ and a male attribute
embedding $c_{2}$, and the occupation would be represented by the
occupation's embedding $w$. Each answer, then, would be given a score

\begin{equation}
\text{score}(\hat{s},c_{i})=(1-\alpha)\text{sim}(c_{1},c_{i})+\alpha\text{sim}(c_{2},c_{i})+\text{sim}(w,c_{i}),\label{eq:score}
\end{equation}
where $\text{sim}(a,b)$ is the cosine similarity between embeddings
$a$ and $b$. The method then picks the answer $c_{i}$ with the
highest score. This score simplifies to 
\[
\text{score}(\hat{s},c_{1})=(1-\alpha)+\alpha\text{sim}(c_{2},c_{1})+\text{sim}(w,c_{1})
\]
and 
\[
\text{score}(\hat{s},c_{2})=(1-\alpha)\text{sim}(c_{1},c_{2})+\alpha+\text{sim}(w,c_{2})
\]
because $\text{sim}(a,a)=1$. To produce the psychometric curve, we
would modify the value of $\alpha$ and obtain several responses.
For the combination of all responses for a particular word $w$, cue
1, and cue 2, we can fit a function to build the psychometric curve
(Eq. \ref{eq:psych-curve-1}). Based on this psychometric curve, then,
we can extract the PSE and JNE. If the PSE is not exactly at $\alpha=0.5$,
we might conclude that the system is biased. If it is between $\alpha\in[0,\frac{1}{2}]$,
we might say that there is a bias against cue 1. An example of several
psychometric curves is in Fig. \ref{fig:Two-kinds-of}.

In our work, the word embeddings model we used is based on skip-gram.
It maps each word into a 100-dimentional continuous vector. If the
input contains multiple words, the embedding is combined by averaging
the embeddings of each word.

\subsection{Estimating distribution of inputs conditioned on outputs}

Using the representation of word embeddings, we can examine the distribution
of embeddings conditioned on classifications that we can be make using
those embeddings. For example, we could compute the posterior distribution
of Glove embeddings using a classifier that predicts sentiments based
on those embeddings. In particular, we train a classifier of positive
(or negative) sentiment $p(+\mid s)$ for an embedding $s$, and are
able to estimate the distribution of embeddings $p(s\mid+)$ conditioned
on positive sentiments. We create the distribution $p(+\mid s)$ using
a multilayer perceptron.

\section{Experiments}

\subsection{Datasets}

We use several datasets with curated labels of gender and other demographic
information. We also use a dataset for training the word embedding
and another dataset for training sentiment analysis.
\begin{description}
\item [{Labor~statistics}] For some analysis, we need to validate our
estimated biases with external data. We use data from labor statistics
on occupations, the number of workers in those occupations, and the
gender breakdown of those workers. The data is based on the U.S. Bureau
of Labor Statistics \cite{bureau2012employed}. This data has been
used before in \cite{caliskan2017semantics} to also externally validate
their method.
\item [{Wikipedia~dataset}] For training the word embedding vectors, we
use a dump of the English Wikipedia dataset from March, 2019.
\item [{Large~Movie~Review}] For training the sentiment analysis predictor,
we use the Large Movie Review Dataset. This is a very popular sentiment
dataset \cite{maas-EtAl:2011:ACL-HLT2011} and it contains movie reviews
from the Internet Movie Database (IMDB) in which reviews with more
than 7 stars (from 1 to 10) get assigned a positive sentiment and
fewer than 4 stars get assigned a negative sentiment.
\item [{Equity~Evaluation~Corpus~(EEC)}] To evaluate biases in sentiment
analysis, we use a dataset of names and relationships associated with
genders. For example, John and uncle are male and Alice and aunt are
female. These relationships are part of the EEC dataset by \cite{DBLP:journals/corr/abs-1805-04508}.
\end{description}

\subsection{Bootstrapping word embedding estimation}

\begin{figure}
\begin{centering}
\includegraphics[scale=0.8]{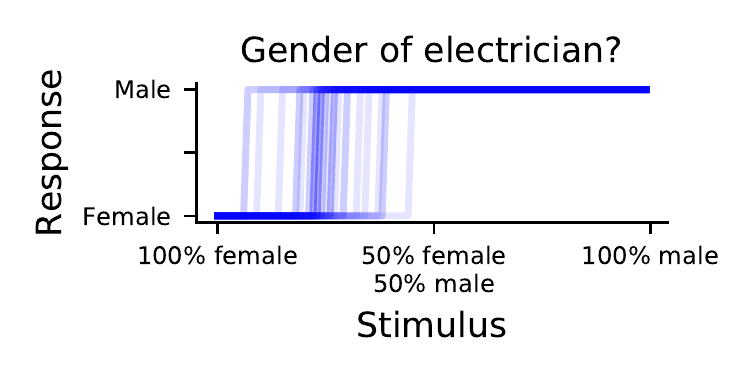}
\par\end{centering}
\caption{\label{fig:Two-kinds-of}Artificial psychometric curves for the \emph{electrician}
occupation. The average point at which it crosses from Female to Male
responses is the Point of Subjective Equivalence (PSE) and the standard
deviation is the Just-noticeable Difference (JND). For this case,
the PSE is to the left of the 50\%/50\% gender stimulus, suggesting
a bias toward male: even though less than 50\% maleness is perceived,
the model thinks that the cue is male.}
\end{figure}

For each word embedding model, we were able to get a PSE from word
pairs given a target word. However, to form a proper psychometric
curve, we need to understand the uncertainty that exists in the model.
We can think of these variations as the noise in the perceptual system
of the AI model, related to $p(s_{p}\mid\hat{s})$ in Eq. \ref{eq:stimulus}.
To estimate this uncertainty, we bootstrapped 32 GloVe word embedding
models. Concretely, we trained word embedding models with the Wikipedia
dataset. Given the size of the dataset, it was unfeasible to perform
direct bootstrap by keeping all data in memory and sampling with replacement.
Instead, we perform a streaming bootstrap repeating each line a random
number of times sampled from a Poisson distribution with mean $\lambda=1$
(see \cite{43157}). After fitting a psychometric curve to these decisions,
if we found any biases in the PSE, the JND would help us understand
how stringent these biases are. High confidence, in this case, would
be represented by low JND. An example of a set of psychometric curves
from this bootstrap process is depicted in Fig. \ref{fig:Two-kinds-of}.

\subsection{Sentiment analysis and sampler}

We use a multi-layer perceptron to learn the probability distribution
of word embeddings to sentiment $p(+\mid s)$. We learn a 100-dimensional
word2vec embedding using the movie review dataset. The dataset consists
of 12500 positive reviews and 12500 negative reviews. We tokenized
reviews, removed symbols, process each reviews using the skip-gram
word embedding model, and generate an embedding by averaging word2vec
vectors for each word. In cross validation, the classifier achieves
a 0.953 AUC score.

For the sampler, we use the Python package \texttt{pyro} to perform
MCMC using a No-U-Turn Sampler \cite{hoffman2014no}. The sampler
used 10,000 warm-up samples and run 10,000 steps after that.

\section{Results}

In this paper, we wanted to adapt the methods developed in experimental
psychology to detect biases in decisions made by artificial intelligence
models. In particular, we adapted the two-alternative forced choice
(2AFC) task to understand biases in word embeddings. We developed
two kinds of experiments: a signal detection theory experiment that
estimates the bias and uncertainty on the bias and a sampling experiment
that estimates the distribution of word embeddings conditioning on
positive sentiment. Both types of experiments provide a window into
how powerful the analogy of psychophysics could be for uncovering
biases in artificial intelligence methods.

\subsection{Measured biases of word embeddings based on 2AFC}

\begin{figure}
\begin{centering}
\subfloat[PSE of occupations. To the left of 50\%/50\%, it is a bias against
female. To the right of 50\%/50\%, it is a bias against male.]{\begin{centering}
\includegraphics{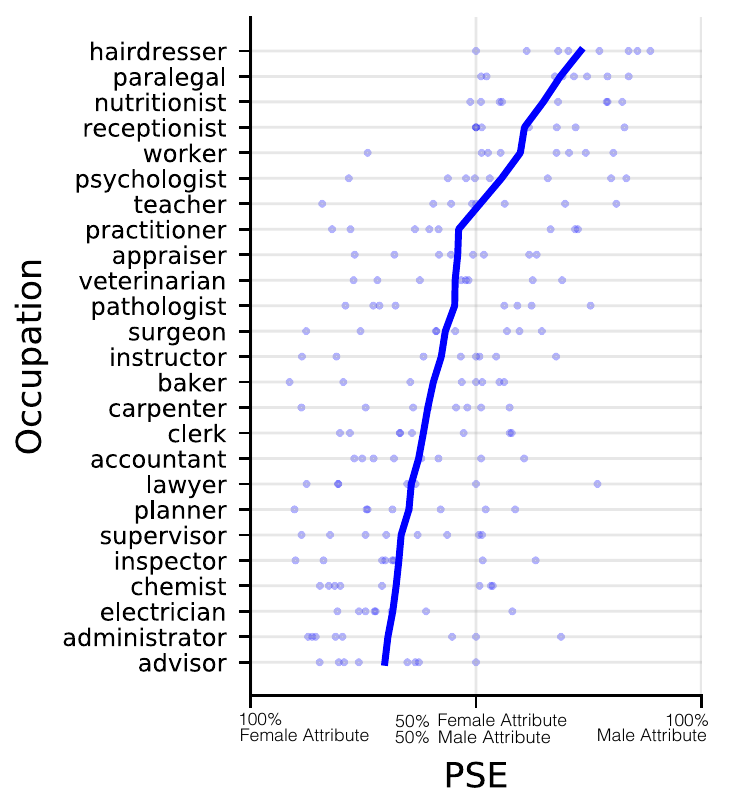}
\par\end{centering}
}
\par\end{centering}
\begin{centering}
\subfloat[JND of occupations. The lower the JND, the higher confidence in the
judgement of bias (PSE)]{\begin{centering}
\includegraphics{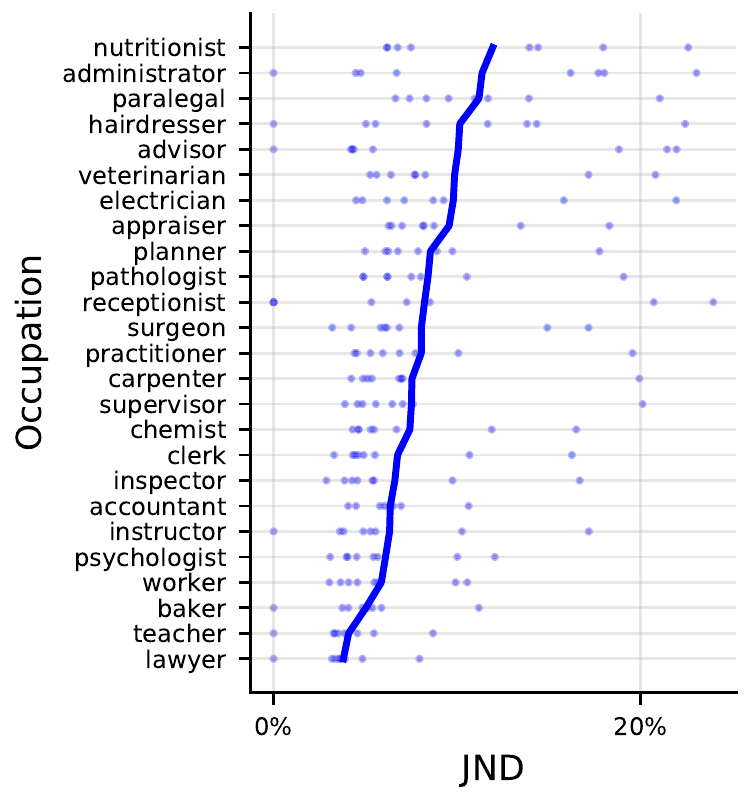}
\par\end{centering}
}
\par\end{centering}
\caption{\label{fig:2AFC-for-several}2AFC results for occupations: Taking
a set of occupations from the US's Labor Statistics dataset reported
in \cite{caliskan2017semantics}, we calculate the PSE for each of
those occupations based on the distances between embeddings of the
occupation and two cues. For each occupation, we were able to get
a PSE which indicates the percentage of \textquotedblleft maleness\textquotedblright{}
or \textquotedblleft femaleness\textquotedblright{} in the questions
when the AI agent could not decide whether an occupation is a male
or female. Each dot represents the PSE or JND of pairs from female/male,
woman/man, girl/boy, sister/brother, she/he her/him, hers/his, and
daughter/son.}
\end{figure}

The results show relatively intuitive trends in occupations. We first
examine whether the psychometric curves for an example occupation
(``electrician'') vs female--male continuum stimuli based on bootstrap
produced sensible results (Fig. \ref{fig:Two-kinds-of}). It indeed
produces a bias against females. A more systematic examination of
the phenomenon for a sample of occupations and a set of gendered attributes
reveals an intuitive pattern (Fig. \ref{fig:2AFC-for-several}a).
Each point in this graph represents one PSE and JND extracted from
a psychometric curve of one pair of female/male attributes out of
the set female/male, woman/man, girl/boy, sister/brother, she/he her/him,
hers/his, and daughter/son (from \cite{caliskan2017semantics}). More
female-perceived occupations have a bias against male, and vice versa.
There are some biases for which the model is more certain about which
can be observed in the JND results (Fig. \ref{fig:2AFC-for-several}b).
For example, hairdresser is a highly biased occupation against man
but with high uncertainty. On the other hand, lawyer is an occupation
relatively biased against woman with significantly lower uncertainty
than hairdresser. It is important to correlate our results with real
datasets that may point to some ground truth. Therefore, we externally
validate the results on a real-world dataset of gender occupations
based on labor statistics. We found that PSE correlates well with
the percent of total occupations held by man within each occupation
$\rho=0.368$ ($p=0.049$) and the JNE correlates well with the standard
deviation of such proportion $\rho=0.401$ ($p=0.031$).

\begin{figure}
\begin{centering}
\includegraphics{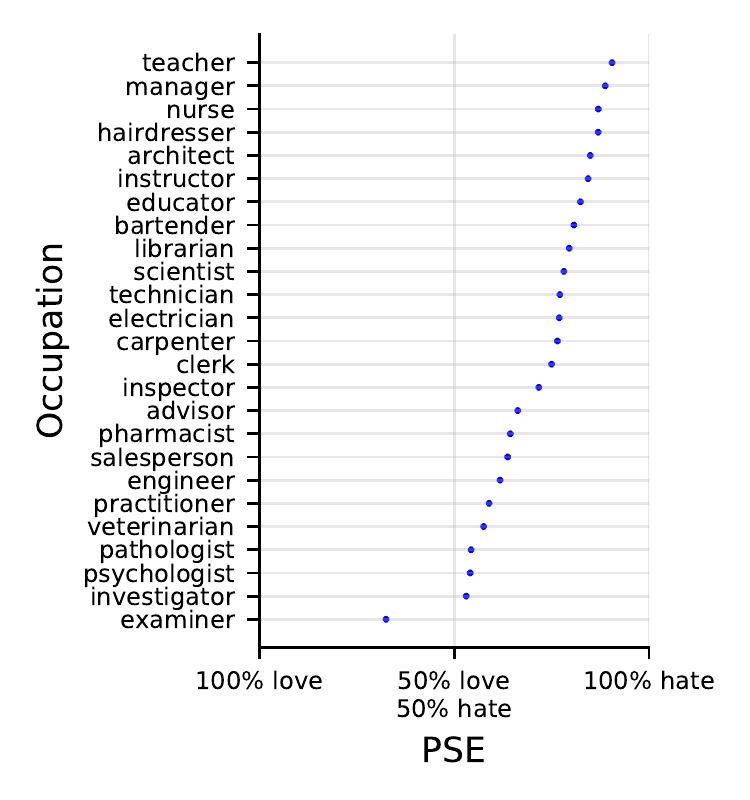}
\par\end{centering}
\caption{\label{fig:2AFC-for-lovehate}PSE of the 2AFC task for occupations
vs love--hate stimulus spectrum. Only one pair of stimuli was tried
on this semantic relationship.}
\end{figure}

We perform a similar analysis to the one above but now choosing as
stimulus that is a combination of love and hate. We found interesting
patterns as well where intuitively more likable occupations such as
teacher have a bias in favor of love whereas (medical) examiner has
a bias in favor of hate (Fig. \ref{fig:2AFC-for-lovehate}a). Additionally,
the bias for teacher is highly confident, as can be observed in the
JND estimations (not shown). For this dataset, however, we do not
have external validation.

\subsection{Measured biases of sentiment analysis predictions based on MCMC}

\begin{figure}
\begin{centering}
\includegraphics[scale=0.8]{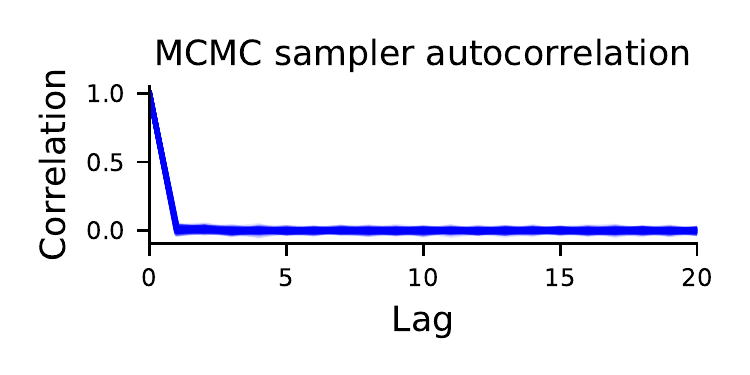}
\par\end{centering}
\caption{\label{fig:Autocorrelation-of-posterior}Autocorrelation of posterior
embeddings conditioned on positive sentiment}

\end{figure}

We first wanted to check that the MCMC sampler has achieved stationarity.
We have used the standard approach of computing autocorrelations of
all dimensions. The sampler has a warmup of 10,000 steps followed
by sample of 10,000 steps. Visual inspection of the autocorrelation
reveals a sharp decline after 1 time lag (Fig. \ref{fig:Autocorrelation-of-posterior}.),
which is a sign that the Markov chain has properly mixed \cite{gilks1995markov}.

\begin{figure}
\centering{}\includegraphics[scale=0.8]{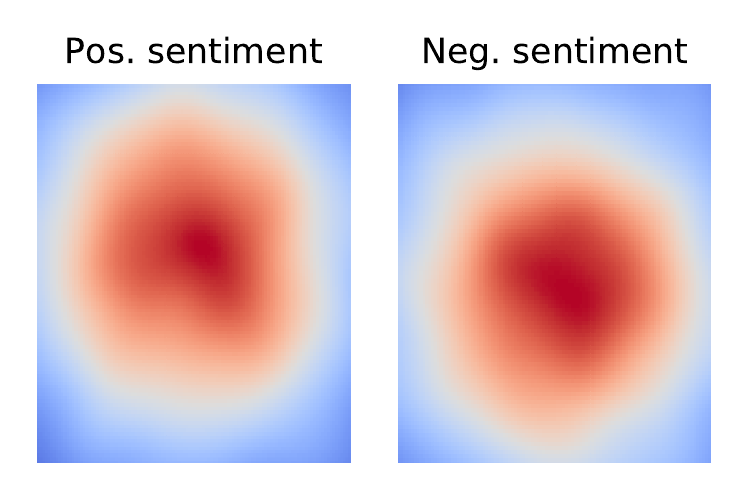}\caption{\label{fig:Posterior-distribution-of}Posterior distribution of embeddings
given positive sentiment responses projected using PCA. Two example
words from a curated dictionary of positive words have intuitive log-likelihood
differences in the posterior distribution.}
\end{figure}

The posterior distribution of embeddings is hard to visualize because
it has 100 dimensions. We perform a dimensionality reduction using
PCA to do so (Fig. \ref{fig:Posterior-distribution-of}). The distributions
of this projection for the posterior conditioned on positive and negative
sentiments are slightly different. This fact could be used to examine
where both distributions differ. However, because we used sentiment
analysis of movie reviews, there is no simple approach to extract
a review from an embedding because our embedding is the average embedding
of each word in the review.

\begin{figure}
\begin{centering}
\includegraphics[scale=0.8]{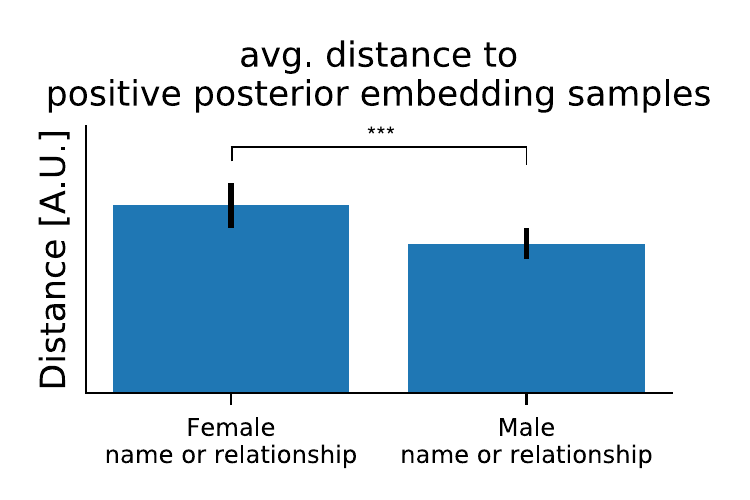}
\par\end{centering}
\caption{\label{fig:Average-distance-of}Average distance of female/male names
or relationships to the posterior distribution of embeddings conditioned
on positive sentiment: Average distance of female/male names or relationships
to the posterior distribution of embeddings conditioned on positive
sentiment. We have a set of names that infer genders from the Equity
Evaluation Corpus. With each name we calculated the distances between
the name and all the sampled embeddings, then we have the average
distances between each word and all the samples. We then calculated
the average distances for both genders.}
\end{figure}

We then wanted to measure whether there are biases in the estimated
distributions. We measured if there are biases in gender from the
posterior distribution. In particular, we measured whether names and
relationships that are usually associated with a gender are closer
or farther to the posterior distribution conditioned on positive sentiment
(Fig. \ref{fig:Average-distance-of}). Indeed, we found that female
names are significantly farther away from this posterior distribution
compared to male names, suggesting that the posterior distribution
has more male-dominated embeddings. However, these results are relatively
minor because we are using movie reviews to learn the embeddings.

\begin{table*}
\subfloat[Closest words, conditioned on positive]{{\scriptsize{}}%
\begin{tabular}{lll}
{\scriptsize{}word} & {\scriptsize{}sentiment} & {\scriptsize{}$\overline{d}$}\tabularnewline
\hline 
{\scriptsize{}wonderful} & \textbf{\scriptsize{}positive} & {\scriptsize{}0.4888}\tabularnewline
{\scriptsize{}unforgettable} & \textbf{\scriptsize{}positive} & {\scriptsize{}0.4895}\tabularnewline
{\scriptsize{}devilish} & {\scriptsize{}negative} & {\scriptsize{}0.4901}\tabularnewline
{\scriptsize{}amazing} & \textbf{\scriptsize{}positive} & {\scriptsize{}0.4907}\tabularnewline
{\scriptsize{}versatility} & \textbf{\scriptsize{}positive} & {\scriptsize{}0.4910}\tabularnewline
{\scriptsize{}heartfelt} & \textbf{\scriptsize{}positive} & {\scriptsize{}0.4911}\tabularnewline
{\scriptsize{}wondrous} & \textbf{\scriptsize{}positive} & {\scriptsize{}0.4911}\tabularnewline
{\scriptsize{}enthusiastically} & \textbf{\scriptsize{}positive} & {\scriptsize{}0.4911}\tabularnewline
{\scriptsize{}cherished} & \textbf{\scriptsize{}positive} & {\scriptsize{}0.4911}\tabularnewline
{\scriptsize{}terrific} & \textbf{\scriptsize{}positive} & {\scriptsize{}0.4912}\tabularnewline
\end{tabular}{\scriptsize\par}

}\quad{}\subfloat[Most distant, conditioned on positive]{{\scriptsize{}}%
\begin{tabular}{lll}
{\scriptsize{}word} & {\scriptsize{}sentiment} & {\scriptsize{}$\overline{d}$}\tabularnewline
\hline 
{\scriptsize{}dismally} & \textbf{\scriptsize{}negative} & {\scriptsize{}0.5114}\tabularnewline
{\scriptsize{}incompetent} & \textbf{\scriptsize{}negative} & {\scriptsize{}0.5116}\tabularnewline
{\scriptsize{}redundant} & \textbf{\scriptsize{}negative} & {\scriptsize{}0.5117}\tabularnewline
{\scriptsize{}hopelessly} & \textbf{\scriptsize{}negative} & {\scriptsize{}0.5117}\tabularnewline
{\scriptsize{}mess} & \textbf{\scriptsize{}negative} & {\scriptsize{}0.5118}\tabularnewline
{\scriptsize{}hideously} & \textbf{\scriptsize{}negative} & {\scriptsize{}0.5118}\tabularnewline
{\scriptsize{}lifeless} & \textbf{\scriptsize{}negative} & {\scriptsize{}0.5119}\tabularnewline
{\scriptsize{}substandard} & \textbf{\scriptsize{}negative} & {\scriptsize{}0.5126}\tabularnewline
{\scriptsize{}smother} & \textbf{\scriptsize{}negative} & {\scriptsize{}0.5129}\tabularnewline
{\scriptsize{}dreadfully} & \textbf{\scriptsize{}negative} & {\scriptsize{}0.5131}\tabularnewline
\end{tabular}{\scriptsize\par}

}\quad{}\subfloat[Closest, conditioned on negative]{{\scriptsize{}}%
\begin{tabular}{lll}
{\scriptsize{}word} & {\scriptsize{}sentiment} & {\scriptsize{}$\overline{d}$}\tabularnewline
\hline 
{\scriptsize{}substandard} & \textbf{\scriptsize{}negative} & {\scriptsize{}0.4754}\tabularnewline
{\scriptsize{}hideously} & \textbf{\scriptsize{}negative} & {\scriptsize{}0.4765}\tabularnewline
{\scriptsize{}dreadfully} & \textbf{\scriptsize{}negative} & {\scriptsize{}0.4768}\tabularnewline
{\scriptsize{}smother} & \textbf{\scriptsize{}negative} & {\scriptsize{}0.4768}\tabularnewline
{\scriptsize{}lifeless} & \textbf{\scriptsize{}negative} & {\scriptsize{}0.4773}\tabularnewline
{\scriptsize{}hopelessly} & \textbf{\scriptsize{}negative} & {\scriptsize{}0.4773}\tabularnewline
{\scriptsize{}mess} & \textbf{\scriptsize{}negative} & {\scriptsize{}0.4777}\tabularnewline
{\scriptsize{}dismally} & \textbf{\scriptsize{}negative} & {\scriptsize{}0.4778}\tabularnewline
{\scriptsize{}redundant} & \textbf{\scriptsize{}negative} & {\scriptsize{}0.4780}\tabularnewline
{\scriptsize{}pointless} & \textbf{\scriptsize{}negative} & {\scriptsize{}0.4780}\tabularnewline
\end{tabular}{\scriptsize\par}

}\quad{}\subfloat[Most distant, conditioned on negative]{{\scriptsize{}}%
\begin{tabular}{lll}
{\scriptsize{}word} & {\scriptsize{}sentiment} & {\scriptsize{}$\overline{d}$}\tabularnewline
\hline 
{\scriptsize{}beautifully} & \textbf{\scriptsize{}positive} & {\scriptsize{}0.5153}\tabularnewline
{\scriptsize{}uncompromising} & {\scriptsize{}negative} & {\scriptsize{}0.5154}\tabularnewline
{\scriptsize{}supremacy} & \textbf{\scriptsize{}positive} & {\scriptsize{}0.5156}\tabularnewline
{\scriptsize{}heartfelt} & \textbf{\scriptsize{}positive} & {\scriptsize{}0.5156}\tabularnewline
{\scriptsize{}devilish} & {\scriptsize{}negative} & {\scriptsize{}0.5162}\tabularnewline
{\scriptsize{}amazing} & \textbf{\scriptsize{}positive} & {\scriptsize{}0.5165}\tabularnewline
{\scriptsize{}versatility} & \textbf{\scriptsize{}positive} & {\scriptsize{}0.5167}\tabularnewline
{\scriptsize{}cherished} & \textbf{\scriptsize{}positive} & {\scriptsize{}0.5170}\tabularnewline
{\scriptsize{}unforgettable} & \textbf{\scriptsize{}positive} & {\scriptsize{}0.5186}\tabularnewline
{\scriptsize{}wonderful} & \textbf{\scriptsize{}positive} & {\scriptsize{}0.5191}\tabularnewline
\end{tabular}{\scriptsize\par}

}

\caption{Average distance of words from curated dictionary of words with sentiment
to posterior embeddings conditioned on positive sentiment and negative
sentiment, ranked by distance}
\end{table*}

We externally validate the sampler using a curated dictionary of sentiments.
We compute the average distance of the embeddings of all these words
to the posterior distributions conditioned on positive and negative
sentiments. We found a negative correlation between distance to the
posterior conditioned on positive sentiment and positive words ($\rho=-0.4$,
$n=6,789$, $p<0.001$). Similarly, we found a positive correlation
between distance to the posterior conditioned on negative sentiment
and positive words. These results suggest that the distance to the
distribution provides not only a predictor to the sentiment of words
but also a natural ordering of those words with respect to the conditioning
of the MCMC sampler.

\section{Discussion}

In this work, we use artificial psychophysics to detect biases in
AI. We show its application to word embeddings and sentiment analysis
predictions. Our method was able to capture similar biases that have
been reported in the literature but using a more coherent and perhaps
simpler set of ideas. However, there are some shortcomings that we
now discuss.

We need a more systematic evaluation of the method. We need to see
if we find similar effects to those found by more specialized techniques
such as IAT. For example, while we find a correlation between the
labor statistics data and the PSE (Pearson\textquoteright s correlation
coefficient $\rho=0.39$), this correlation was not as strong as the
one found through the IAT task in \cite{caliskan2017semantics} (Pearson\textquoteright s
correlation coefficient $\rho=0.9$). One disadvantage of IAT is that
it needs a basket of words to represent the attributes that one wants
to analyze. For example, while our method only needs the embedding
of ``woman'' for one option and the embedding of ``man'' for the
other, IAT used a set of female names and a set of male names. We
could easily extend our technique to include all pairwise PSEs and
JNDs that then would be average and could perhaps improve the correlation.
However, this seems unlikely given the large dispersion of values
in the current estimations. We will explore this approach in the future.

The two-alternative forced choice task (2AFC) has limitations that
also apply to our task. For one, it can only handle \emph{two} alternatives
at the same time which makes it inefficient to explore multiple, simultaneous
relationships. A possible fix to this issue is to fit several pairwise
psychometric curves but the interpretation becomes significantly more
cumbersome \cite{green1966signal}. However, we believe that the rich
history and theoretical foundation of the method outweigh the issues
of multiple comparisons. If anything, our use of MCMC can do multiple,
simultaneous examinations of the underlying method but the way to
apply is not as straightforward as the 2AFC.

As with all the methods in this area, the evaluation of our results
is difficult. While we have a dataset from labor statistics that relates
to the PSE of occupation vs female--male attributes, if we did not
find such relationship, the bias would still be there---high false
negatives. This is apparent for the occupation vs love--hate experience.
We found intuitive relationships between lovable and undesirable occupations
(e.g., teacher being the most lovable and (medical) examiner being
the most hated) but so far we do not have a validation set for it.
In the future, we will attempt to validate these results using survey
information, such as the Pew Research Center survey on trust of different
occupations \cite{funk2017public}. Similarly, checking the posterior
distribution of the MCMC result is perhaps even more challenging.
In a sense, we need a way to generate interpretable data from a point
in the posterior distribution. For example, in our sentiment analysis
predictor, the posterior is on the embedding space, making it almost
impossible to map the embedding into a movie review. While this seems
that defeats the purpose of the posterior, we are still able to capture
some trends in the data whereas the embeddings of positive words are
intuitively closer to the posterior distribution conditioned on positive
sentiment. In the future, we will explore representations that are
more interpretable and that therefore will allow to examine the posterior
more easily. One obvious experiment to try is an embedding that involves
images.

The biases we are detecting do not necessarily constitute a problematic
feature of an AI system that is attempting to make the most accurate
predictions. After all, biases in the statistical sense can help a
system prevent overfitting, and a great deal of modern machine learning
techniques uses an array of methods for introducing biases explicitly---e.g.,
regularization, dropout, and data augmentation can all be seen as
increasing bias \cite{Goodfellow-et-al-2016}. Also, humans themselves
seem to incorporate biases in an optimal manner in the form of a-priori
knowledge \cite{tenenbaum2011grow}. However, these kinds of biases
are best understood in supervised learning scenarios where there is
a clear measure of performance \cite{hastie2005elements}. As such,
it seems problematic that systems that are unsupervised, such as word
embeddings, contain biases on attributes that are protected. Companies
and the public seem to agree: anecdotally, almost a year ago, we were
able to reproduce biases using the built-in word embedding in the
Python's software package spaCy \cite{spacy2}. When we re-attempted
such experiment with a more recent version of the software, however,
we were not able find such biases. This suggests that software companies
and the general machine learning community recognizes that these types
of biases might be unacceptable (e.g., \cite{buolamwini2018gender}).

One of the ideal goals of this work is to not only detect biases but
fix them. There are many proposals to fix biases in AI models but
to the best of our knowledge we are not aware of debiasing methods
based on experimental psychology or psychophysics. Perhaps building
a system that fixes these issues would greatly inform how we design
our detection task. Maybe some of the biases that we detect are not
fixable or biases that we think are hard to detect are easily fixable.
We will explore this interplay in the future.

We believe that our proposal can open the door to collaboration across
disciplines. For example, there is rich literature on quantitative
methods for cognitive-behavior therapy for cognitive debiasing \cite{croskerry2013cognitive}.
More interestingly, perhaps the methods other researchers have developed
for detecting and fixing biases in AI systems can be \emph{transported
back} to cognitive-behavioral therapy.

\section{Conclusion}

In this work, we proposed a method for detecting biases in AI models
using a coherent intellectual framework rooted in Experimental Psychology
and Psychophysics. We adapted the alternative forced choice (2AFC)
task and a sampling mechanism based on MCMC to examine these biases.
We evaluated gender biases in a word embedding model trained on Wikipedia
and a sentiment analysis model trained on movie reviews. Our results
suggest that we are able to detect these biases while keeping a conceptual
language that is common to what is used in Psychophysics.

In the future, we will explore how to adapt other ideas from experimental
psychology to detect and even fix issues found in AI models. We believe
that many of the issues found in AI can be fixed effectively without
significant loss of performance. Also, we believe that akin to how
humans who have been subjected to racist, sexist, and extremist views
can be rehabilitated through deradicalization and disengagement \cite{stern2010deradicalization},
AI models can also be rehabilitated.

\section*{Acknowledgements}

L. Liang and D. E. Acuna were partially funded by NSF grant \#1800956
and ORI grant ``Methods and tools for scalable figure reuse detection
with statistical certainty reporting''. The authors would like to
thank Xinxuan Wei for preliminary discussions and contributions to
the work.

\bibliographystyle{IEEEtran}
\bibliography{AI_Bias}

\begin{thebibliography}{10}
\providecommand{\url}[1]{#1}
\csname url@samestyle\endcsname
\providecommand{\newblock}{\relax}
\providecommand{\bibinfo}[2]{#2}
\providecommand{\BIBentrySTDinterwordspacing}{\spaceskip=0pt\relax}
\providecommand{\BIBentryALTinterwordstretchfactor}{4}
\providecommand{\BIBentryALTinterwordspacing}{\spaceskip=\fontdimen2\font plus
\BIBentryALTinterwordstretchfactor\fontdimen3\font minus
  \fontdimen4\font\relax}
\providecommand{\BIBforeignlanguage}[2]{{%
\expandafter\ifx\csname l@#1\endcsname\relax
\typeout{** WARNING: IEEEtran.bst: No hyphenation pattern has been}%
\typeout{** loaded for the language `#1'. Using the pattern for}%
\typeout{** the default language instead.}%
\else
\language=\csname l@#1\endcsname
\fi
#2}}
\providecommand{\BIBdecl}{\relax}
\BIBdecl

\bibitem{stone2016one}
P.~Stone, R.~Brooks, E.~Brynjolfsson, R.~Calo, O.~Etzioni, G.~Hager, and
  A.~Teller, ``One hundred year study on artificial intelligence,''
  \emph{Artificial Intelligence and Life in}, vol. 2030, 2016.

\bibitem{dastin2018amazon}
J.~Dastin, ``Amazon scraps secret ai recruiting tool that showed bias against
  women,'' \emph{San Fransico, CA: Reuters. Retrieved on October}, vol.~9, p.
  2018, 2018.

\bibitem{angwin2016machine}
J.~Angwin, J.~Larson, S.~Mattu, and L.~Kirchner, ``Machine bias,''
  \emph{ProPublica, May}, vol.~23, p. 2016, 2016.

\bibitem{hurley2016credit}
M.~Hurley and J.~Adebayo, ``Credit scoring in the era of big data,'' \emph{Yale
  JL \& Tech.}, vol.~18, p. 148, 2016.

\bibitem{stock2017convnets}
P.~Stock and M.~Cisse, ``Convnets and imagenet beyond accuracy: Explanations,
  bias detection, adversarial examples and model criticism,'' \emph{arXiv
  preprint arXiv:1711.11443}, 2017.

\bibitem{chen2018my}
I.~Chen, F.~D. Johansson, and D.~Sontag, ``Why is my classifier
  discriminatory?'' in \emph{Advances in Neural Information Processing
  Systems}, 2018, pp. 3539--3550.

\bibitem{celis2019classification}
L.~E. Celis, L.~Huang, V.~Keswani, and N.~K. Vishnoi, ``Classification with
  fairness constraints: A meta-algorithm with provable guarantees,'' in
  \emph{Proceedings of the Conference on Fairness, Accountability, and
  Transparency}.\hskip 1em plus 0.5em minus 0.4em\relax ACM, 2019, pp.
  319--328.

\bibitem{fechner}
G.~T. Fechner, \emph{Elemente der Psychophysik (Elements of Psychophysics)},
  1960.

\bibitem{hastie2005elements}
T.~Hastie, R.~Tibshirani, J.~Friedman, and J.~Franklin, \emph{The elements of
  statistical learning: data mining, inference and prediction}.\hskip 1em plus
  0.5em minus 0.4em\relax Springer, 2005.

\bibitem{tan2019efficientnet}
M.~Tan and Q.~V. Le, ``Efficientnet: Rethinking model scaling for convolutional
  neural networks,'' \emph{arXiv preprint arXiv:1905.11946}, 2019.

\bibitem{o2016weapons}
C.~O'Neil, \emph{Weapons of math destruction: How big data increases inequality
  and threatens democracy}.\hskip 1em plus 0.5em minus 0.4em\relax Broadway
  Books, 2016.

\bibitem{dovsilovic2018explainable}
F.~K. Do{\v{s}}ilovi{\'c}, M.~Br{\v{c}}i{\'c}, and N.~Hlupi{\'c}, ``Explainable
  artificial intelligence: A survey,'' in \emph{2018 41st International
  convention on information and communication technology, electronics and
  microelectronics (MIPRO)}.\hskip 1em plus 0.5em minus 0.4em\relax IEEE, 2018,
  pp. 0210--0215.

\bibitem{sarkar2016accuracy}
S.~Sarkar, T.~Weyde, A.~Garcez, G.~G. Slabaugh, S.~Dragicevic, and C.~Percy,
  ``Accuracy and interpretability trade-offs in machine learning applied to
  safer gambling,'' in \emph{CEUR Workshop Proceedings}, vol. 1773.\hskip 1em
  plus 0.5em minus 0.4em\relax CEUR Workshop Proceedings, 2016.

\bibitem{domingos2015master}
P.~Domingos, \emph{The master algorithm: How the quest for the ultimate
  learning machine will remake our world}.\hskip 1em plus 0.5em minus
  0.4em\relax Basic Books, 2015.

\bibitem{mcduff2019characterizing}
D.~McDuff, S.~Ma, Y.~Song, and A.~Kapoor, ``Characterizing bias in classifiers
  using generative models,'' \emph{arXiv preprint arXiv:1906.11891}, 2019.

\bibitem{caliskan2017semantics}
A.~Caliskan, J.~J. Bryson, and A.~Narayanan, ``Semantics derived automatically
  from language corpora contain human-like biases,'' \emph{Science}, vol. 356,
  no. 6334, pp. 183--186, 2017.

\bibitem{leibo2018psychlab}
J.~Z. Leibo, C.~d.~M. d'Autume, D.~Zoran, D.~Amos, C.~Beattie, K.~Anderson,
  A.~G. Casta{\~n}eda, M.~Sanchez, S.~Green, A.~Gruslys \emph{et~al.},
  ``Psychlab: a psychology laboratory for deep reinforcement learning agents,''
  \emph{arXiv preprint arXiv:1801.08116}, 2018.

\bibitem{green1966signal}
D.~M. Green, J.~A. Swets \emph{et~al.}, \emph{Signal detection theory and
  psychophysics}.\hskip 1em plus 0.5em minus 0.4em\relax Wiley New York, 1966,
  vol.~1.

\bibitem{acuna2015using}
D.~E. Acuna, M.~Berniker, H.~L. Fernandes, and K.~P. Kording, ``Using
  psychophysics to ask if the brain samples or maximizes,'' \emph{Journal of
  vision}, vol.~15, no.~3, pp. 7--7, 2015.

\bibitem{charpentier1891analyse}
A.~Charpentier, ``Analyse experimentale de quelques elements de la sensation de
  poids,'' \emph{Archive de Physiologie normale et pathologiques}, vol.~3, pp.
  122--135, 1891.

\bibitem{harmening2014mapping}
W.~M. Harmening, W.~S. Tuten, A.~Roorda, and L.~C. Sincich, ``Mapping the
  perceptual grain of the human retina,'' \emph{Journal of Neuroscience},
  vol.~34, no.~16, pp. 5667--5677, 2014.

\bibitem{bornstein2018antidiscriminatory}
S.~Bornstein, ``Antidiscriminatory algorithms,'' \emph{Ala. L. Rev.}, vol.~70,
  p. 519, 2018.

\bibitem{benthall2019racial}
S.~Benthall and B.~D. Haynes, ``Racial categories in machine learning,'' in
  \emph{Proceedings of the Conference on Fairness, Accountability, and
  Transparency}.\hskip 1em plus 0.5em minus 0.4em\relax ACM, 2019, pp.
  289--298.

\bibitem{murphy2012machine}
K.~P. Murphy, \emph{Machine learning: a probabilistic perspective}.\hskip 1em
  plus 0.5em minus 0.4em\relax MIT press, 2012.

\bibitem{pearl2009causality}
J.~Pearl, \emph{Causality}.\hskip 1em plus 0.5em minus 0.4em\relax Cambridge
  university press, 2009.

\bibitem{imbens2015causal}
G.~W. Imbens and D.~B. Rubin, \emph{Causal inference in statistics, social, and
  biomedical sciences}.\hskip 1em plus 0.5em minus 0.4em\relax Cambridge
  University Press, 2015.

\bibitem{kusner2017counterfactual}
M.~J. Kusner, J.~Loftus, C.~Russell, and R.~Silva, ``Counterfactual fairness,''
  in \emph{Advances in Neural Information Processing Systems}, 2017, pp.
  4066--4076.

\bibitem{kilbertus2019sensitivity}
N.~Kilbertus, P.~J. Ball, M.~J. Kusner, A.~Weller, and R.~Silva, ``The
  sensitivity of counterfactual fairness to unmeasured confounding,''
  \emph{arXiv preprint arXiv:1907.01040}, 2019.

\bibitem{wu2019counterfactual}
Y.~Wu, L.~Zhang, and X.~Wu, ``Counterfactual fairness: Unidentification, bound
  and algorithm,'' in \emph{Proceedings of the Twenty-Eighth International
  Joint Conference on Artificial Intelligence, IJCAI}, 2019, pp. 10--16.

\bibitem{murray1993perspective}
D.~J. Murray, ``A perspective for viewing the history of psychophysics,''
  \emph{Behavioral and Brain Sciences}, vol.~16, no.~1, pp. 115--137, 1993.

\bibitem{berniker2010learning}
M.~Berniker, M.~Voss, and K.~Kording, ``Learning priors for bayesian
  computations in the nervous system,'' \emph{PloS one}, vol.~5, no.~9, p.
  e12686, 2010.

\bibitem{kording2004bayesian}
K.~P. K{\"o}rding and D.~M. Wolpert, ``Bayesian integration in sensorimotor
  learning,'' \emph{Nature}, vol. 427, no. 6971, p. 244, 2004.

\bibitem{ernst2002humans}
M.~O. Ernst and M.~S. Banks, ``Humans integrate visual and haptic information
  in a statistically optimal fashion,'' \emph{Nature}, vol. 415, no. 6870, p.
  429, 2002.

\bibitem{yuille1993bayesian}
A.~L. Yuille and H.~H. B{\"u}lthoff, ``Bayesian decision theory and
  psychophysics,'' 1993.

\bibitem{knill1996perception}
D.~C. Knill and W.~Richards, \emph{Perception as Bayesian inference}.\hskip 1em
  plus 0.5em minus 0.4em\relax Cambridge University Press, 1996.

\bibitem{sanborn2010uncovering}
A.~N. Sanborn, T.~L. Griffiths, and R.~M. Shiffrin, ``Uncovering mental
  representations with markov chain monte carlo,'' \emph{Cognitive psychology},
  vol.~60, no.~2, pp. 63--106, 2010.

\bibitem{hoffman2014no}
M.~D. Hoffman and A.~Gelman, ``The no-u-turn sampler: adaptively setting path
  lengths in hamiltonian monte carlo.'' \emph{Journal of Machine Learning
  Research}, vol.~15, no.~1, pp. 1593--1623, 2014.

\bibitem{gilks1995markov}
W.~R. Gilks, S.~Richardson, and D.~Spiegelhalter, \emph{Markov chain Monte
  Carlo in practice}.\hskip 1em plus 0.5em minus 0.4em\relax Chapman and
  Hall/CRC, 1995.

\bibitem{schnabel2015evaluation}
T.~Schnabel, I.~Labutov, D.~Mimno, and T.~Joachims, ``Evaluation methods for
  unsupervised word embeddings,'' in \emph{Proceedings of the 2015 Conference
  on Empirical Methods in Natural Language Processing}, 2015, pp. 298--307.

\bibitem{pennington2014glove}
J.~Pennington, R.~Socher, and C.~Manning, ``Glove: Global vectors for word
  representation,'' in \emph{Proceedings of the 2014 conference on empirical
  methods in natural language processing (EMNLP)}, 2014, pp. 1532--1543.

\bibitem{bureau2012employed}
B.~of~Labor~Statistics, ``Employed persons by detailed occupation, sex, race,
  and hispanic or latino ethnicity,'' 2012.

\bibitem{maas-EtAl:2011:ACL-HLT2011}
\BIBentryALTinterwordspacing
A.~L. Maas, R.~E. Daly, P.~T. Pham, D.~Huang, A.~Y. Ng, and C.~Potts,
  ``Learning word vectors for sentiment analysis,'' in \emph{Proceedings of the
  49th Annual Meeting of the Association for Computational Linguistics: Human
  Language Technologies}.\hskip 1em plus 0.5em minus 0.4em\relax Portland,
  Oregon, USA: Association for Computational Linguistics, June 2011, pp.
  142--150. [Online]. Available: \url{http://www.aclweb.org/anthology/P11-1015}
\BIBentrySTDinterwordspacing

\bibitem{DBLP:journals/corr/abs-1805-04508}
\BIBentryALTinterwordspacing
S.~Kiritchenko and S.~M. Mohammad, ``Examining gender and race bias in two
  hundred sentiment analysis systems,'' \emph{CoRR}, vol. abs/1805.04508, 2018.
  [Online]. Available: \url{http://arxiv.org/abs/1805.04508}
\BIBentrySTDinterwordspacing

\bibitem{43157}
N.~Chamandy, O.~Muralidharan, A.~Najmi, and S.~Naidu, ``Estimating uncertainty
  for massive data streams,'' Google, Tech. Rep., 2012.

\bibitem{funk2017public}
C.~Funk and B.~Kennedy, ``Public confidence in scientists has remained stable
  for decades,'' \emph{Pew Research Center}, 2017.

\bibitem{Goodfellow-et-al-2016}
I.~Goodfellow, Y.~Bengio, and A.~Courville, \emph{Deep Learning}.\hskip 1em
  plus 0.5em minus 0.4em\relax MIT Press, 2016,
  \url{http://www.deeplearningbook.org}.

\bibitem{tenenbaum2011grow}
J.~B. Tenenbaum, C.~Kemp, T.~L. Griffiths, and N.~D. Goodman, ``How to grow a
  mind: Statistics, structure, and abstraction,'' \emph{science}, vol. 331, no.
  6022, pp. 1279--1285, 2011.

\bibitem{spacy2}
M.~Honnibal and I.~Montani, ``{spaCy 2}: Natural language understanding with
  {B}loom embeddings, convolutional neural networks and incremental parsing,''
  2017, to appear.

\bibitem{buolamwini2018gender}
J.~Buolamwini and T.~Gebru, ``Gender shades: Intersectional accuracy
  disparities in commercial gender classification,'' in \emph{Conference on
  fairness, accountability and transparency}, 2018, pp. 77--91.

\bibitem{croskerry2013cognitive}
P.~Croskerry, G.~Singhal, and S.~Mamede, ``Cognitive debiasing 1: origins of
  bias and theory of debiasing,'' \emph{BMJ Qual Saf}, vol.~22, no. Suppl 2,
  pp. ii58--ii64, 2013.

\bibitem{stern2010deradicalization}
J.~Stern, ``Deradicalization or disengagement of terrorists: Is it possible,''
  \emph{Future Challenges in National Security and Law}, pp. 1--18, 2010.

\end{thebibliography}

\end{document}